\documentclass[runningheads]{llncs}
\usepackage[usenames, dvipsnames]{color}
\usepackage{amssymb}
\usepackage{graphicx}
\usepackage{subfigure}
\usepackage{xspace}
\usepackage{caption}
\usepackage{algorithm}
\usepackage{algpseudocode}
\usepackage{amsmath}
\usepackage[
  separate-uncertainty = true,
  multi-part-units = repeat
]{siunitx}

\usepackage[colorlinks=true,bookmarks=false]{hyperref}
\usepackage{url}

\newcommand{\true}{\texttt{true}}

\newcommand{\until}{{\textbf{U}}\xspace}
\newcommand{\nextstate}{{\textbf{X}}\xspace}

\newcommand{\comment}[1]{}

\newcommand{\commentout}[1]{}

\begin{document}

\title{Safety-Aware Apprenticeship Learning}
\author{Weichao Zhou \and Wenchao Li}
\authorrunning{Safety-Aware Apprenticeship Learning}
\institute{Department of Electrical and Computer Engineering\\
Boston University\\
\{zwc662, wenchao\}@bu.edu}
\toctitle{Safety-Aware Apprenticeship Learning}
\maketitle

\begin{abstract}
Apprenticeship learning (AL) is a kind of Learning from Demonstration techniques where the reward function of a Markov Decision Process (MDP) is unknown to the learning agent and the agent has to derive a good policy by observing an expert's demonstrations. In this paper, we study the problem of how to make AL algorithms inherently safe while still meeting its learning objective. We consider a setting where the unknown reward function is assumed to be a linear combination of a set of state features, and the safety property is specified in Probabilistic Computation Tree Logic (PCTL). By embedding probabilistic model checking inside AL, we propose a novel {\it counterexample-guided} approach that can ensure safety while retaining performance of the learnt policy. We demonstrate the effectiveness of our approach on several challenging AL scenarios where safety is essential.
\end{abstract}

\section{Introduction}
\label{sec:intro}
The rapid progress of artficial intelligence (AI) comes with a growing concern over its safety when deployed in real-life systems and situations. 
As highlighted in \cite{AmodeiOSCSM16}, if the objective function of an AI agent is wrongly specified, then maximizing that objective function may lead to harmful results. 
In addition, the objective function or the training data may focus only on accomplishing a specific task and ignore other aspects, such as safety constraints, of the environment. 
In this paper, we propose a novel framework that combines explicit safety specification with learning from data. 
We consider safety specification expressed in Probabilistic Computation Tree Logic (PCTL) and 
show how probabilistic model checking can be used to ensure safety and retain performance of a learning algorithm known as {\it apprenticeship learning} (AL).

We consider the formulation of apprenticeship learning by Abbeel and Ng~\cite{Abbeel:2004:ALV:1015330.1015430}.
The concept of AL is closely related to {\it reinforcement learning} (RL) where an agent learns what actions to take in an environment (known as a policy) by maximizing some notion of long-term reward. In AL, however, the agent is not given the reward function, but instead has to first estimate it from a set of expert demonstrations via a technique called {\it inverse reinforcement learning}~\cite{Ng:2000:AIR:645529.657801} . The formulation assumes that the reward function is expressible as a linear combination of {\it known state features}. An expert demonstrates the task by maximizing this reward function and the agent tries to derive a policy that can match the feature expectations of the expert's demonstrations.
Apprenticeship learning can also be viewed as an instance of the class of techniques known as Learning from Demonstration (LfD). 
One issue with LfD is that {\it the expert often can only demonstrate how the task works but not how the task may fail}. 
This is because failure may cause irrecoverable damages to the system such as crashing a vehicle. 
In general, the lack of ``negative examples" can cause a heavy bias in how the learning agent constructs the reward estimate.
In fact, {\it even if all the demonstrations are safe, the agent may still end up learning an unsafe policy}.

The key idea of this paper is to incorporate formal verification in apprenticeship learning. 
We are inspired by the line of work on formal inductive synthesis~\cite{jha-ai2017} and counterexample-guided inductive synthesis~\cite{CEGIS}. 
Our approach is also similar in spirit to the recent work on safety-constrained reinforcement learning~\cite{junges2016safety}. 
However, our approach uses the results of model checking in a novel way. 
We consider safety specification expressed in probabilistic computation tree logic (PCTL). 
We employ a verification-in-the-loop approach by embedding PCTL model checking as a safety checking mechanism inside the learning phase of AL. 
In particular, when a learnt policy does not satisfy the PCTL formula, we leverage counterexamples generated by the model checker to steer the policy search in AL. 
In essence, counterexample generation can be viewed as supplementing negative examples for the learner. 
Thus, the learner will try to find a policy that not only imitates the expert's demonstrations but also stays away from the failure scenarios as captured by the counterexamples. 

In summary, we make the following contributions in this paper. 
\begin{itemize}
\item We propose a novel framework for incorporating formal safety guarantees in Learning from Demonstration.
\item We develop a novel algorithm called \underline{C}ounter\underline{E}xample \underline{G}uided \underline{A}pprenticeship \underline{L}earning (CEGAL) that combines probabilistic model checking with the optimization-based approach of apprenticeship learning. 
\item We demonstrate that our approach can guarantee safety for a set of case studies and attain performance comparable to that of using apprenticeship learning alone.
\end{itemize}

The rest of the paper is organized as follows. 
Section~\ref{sec:prelim} reviews background information on apprenticeship learning and PCTL model checking. 
Section~\ref{sec:overview} defines the safety-aware apprenticeship learning problem and gives an overview of our approach. 
Section~\ref{sec:counterexample} illustrates the counterexample-guided learning framework. 
Section~\ref{sec:safeAL} describes the proposed algorithm in detail. 
Section~\ref{sec:exp} presents a set of experimental results demonstrating the effectiveness of our approach. 
Section~\ref{sec:related} discusses related work. 
Section~\ref{sec:conclu} concludes and offers future directions.

\section{Preliminaries}
\label{sec:prelim}
\graphicspath{{./}}

\subsection{Markov Decision Process and Discrete-Time Markov Chain}
\label{sec:mdp-prelim}

Markov Decision Process (MDP) is a tuple $M = (S,$ $A,$ $T,$ $\gamma,$ $s_0,$ $R)$, where $S$ is a finite set of states; $A$ is a set of actions; $T: S\times A\times S\rightarrow [0, 1]$ is a transition function describing the probability of transitioning from one state $s\in S$ to another state by taking action $a\in A$ in state $s$; $R: S\to \mathbb{R}\ $is a reward function which maps each state $s\in S$ to a real number indicating the reward of being in state $s$; $s_0\in S$ is the initial state; $\gamma \in [0, 1)$ is a discount factor which describes how future rewards attenuate when a sequence of transitions is made. A deterministic and stationary (or memoryless) policy $\pi: S \to A$ for an MDP $M$ is a mapping from states to actions, i.e. the policy deterministically selects what action to take solely based on the current state. In this paper, we restrict ourselves to deterministic and stationary policy. 
A policy $\pi$ for an MDP $M$ induces a Discrete-Time Markov Chain (DTMC) $M_{\pi}=(S, T_\pi, s_0)$, where $T_\pi:S\times S\to [0, 1]$ is the probability of transitioning from a state $s$ to another state in one step. A trajectory $\tau = s_0\xrightarrow{T_\pi(s_0, s_1)>0} s_1\xrightarrow{T_\pi(s_1, s_2)>0} s_2, ...$, is a sequence of states where $s_i\in S$. The accumulated reward of $\tau$ is $\sum\limits\limits_{i=0}^{\infty} \gamma^i R(s_i)$. 
The value function $V_\pi: S\to \mathbb{R}$ measures the expectation of accumulated reward $E[\sum\limits_{i=0}^{\infty}\gamma^i R(s_i)]$ starting from a state $s$ and following policy $\pi$. 
An {\it optimal policy} $\pi$ for MDP $M$ is a policy that maximizes the value function ~\cite{bellman}.

\subsection{Apprenticeship Learning via Inverse Reinforcement Learning}
\label{sec:al-prelim}
{\it Inverse reinforcement learning (IRL)} aims at recovering the reward function $R$ of $M\backslash R = (S, A, T, \gamma, s_0)$
from a set of $m$ trajectories $\Gamma_E=\{\tau_0, \tau_1, ..., \tau_{m-1}\}$ demonstrated by an expert.
{\it Apprenticeship learning (AL)}~\cite{Abbeel:2004:ALV:1015330.1015430} assumes that the reward function is a linear combination of state features, 
i.e.  $R(s) = \omega^Tf(s)$ where $f : S \to [0,\ 1]^k$ is a vector of known features over states $S$ and 
$\omega \in \mathbb{R}^k$ is an unknown weight vector that satisfies $||\omega||_2\leq 1$. 
The expected features of a policy $\pi$ are the expected values of the cumulative discounted state features $f(s)$ by following $\pi$ on $M$, i.e. $\mu_\pi = E[\sum^\infty_{t=0} \gamma^t f(s_t) | \pi]$.
Let $\mu_{E}$ denote the expected features of the unknown expert's policy $\pi_E$. $\mu_{E}$ can be approximated by the expected features of expert's $m$ demonstrated trajectories $\hat{\mu}_E=\frac{1}{m} \sum\limits_{\tau\in\Gamma_E}\sum\limits_{t=0}^{\infty}\gamma^t f(s_{t})$ if $m$ is large enough. With a slight abuse of notations, 
we use $\mu_\Gamma$ to also denote the expected features of a set of paths $\Gamma$.
Given an error bound $\epsilon$, a policy $\pi^*$ is defined to be \textit{$\epsilon$-close} to $\pi_E$ if its expected features $\mu_{\pi^*}$ satisfies $||\mu_E - \mu_{\pi^*}||_2 \leq \epsilon$. 
The expected features of a policy can be calculated by using Monte Carlo method,
value iteration or linear programming~\cite{Abbeel:2004:ALV:1015330.1015430,bellman}.   

The algorithm proposed by Abbeel and Ng~\cite{Abbeel:2004:ALV:1015330.1015430} starts with a random policy $\pi_0$ and its expected features $\mu_{\pi_0}$. Assuming that in iteration $i$, a set of $i$ candidate policies $\Pi=\{\pi_0, \pi_1, ..., \pi_{i-1}\}$ and their corresponding expected features $\{\mu_\pi|\pi\in\Pi\}$ have been found, the algorithm solves the following optimization problem.

\vspace{-1mm}
\begin{equation}
\delta = \max\limits_{\omega}\min\limits_{\pi\in\Pi}\ \omega^T(\hat{\mu}_E - \mu_{\pi})\qquad s.t.\:||\omega||_2\leq 1 
\label{eq:sec1_1}
\end{equation}
\vspace{-1mm}

The optimal $\omega$ is used to find the corresponding optimal policy $\pi_{i}$ and the expected features $\mu_{\pi_i}$. If $\delta\leq\epsilon$, then the algorithm terminates and $\pi_{i}$ is produced as the output. Otherwise, $\mu_{\pi_i}$ is added to the set of features for the candidate policy set $\Pi$ and the algorithm continues to the next iteration.               

\subsection{PCTL Model Checking}

Probabilistic model checking can be used to verify properties of a stochastic system such as ``is the probability that the agent reaches the unsafe area within 10 steps smaller than 5\%?''. \emph{Probabilistic Computation Tree Logic} (PCTL)~\cite{Hansson1994} allows for probabilistic quantification of properties. The syntax of PCTL includes state formulas and path formulas~\cite{kwiatkowska2002prism}. A state formula $\phi$ asserts property of a single state $s\in S$ whereas a path formula $\psi$ asserts property of a trajectory. 

\vspace{-2mm}
\begin{eqnarray}
\centering
\phi &::=& true\ |\ l_i\ |\ \neg \phi_i\ |\ \phi_i \land \phi_j\ |\ P_{\Join p^*}[\psi]\label{eq:sec2_statef}\\
\psi &::=& \nextstate\phi\ |\ \phi_1 \until^{\leq k}\phi_2\ |\ \phi_1\until\phi_2\ 
\end{eqnarray}
\vspace{-2mm}

\noindent where $l_i$ is atomic proposition and $\phi_i, \phi_j$ are state formulas; $\Join\in\{\leq, \geq, <, >\}$; $P_{\Join p^*}[\psi]$ means that the probability of generating a trajectory that satisfies formula $\psi$ is $\Join p^*$. $\nextstate \phi$ asserts that the next state after initial state in the trajectory satisfies $\phi$; $\phi_1 \until^{\leq k}\phi_2$ asserts that $\phi_2$ is satisfied in at most $k$ transitions and all preceding states satisfy $\phi_1$; $\phi_1 \until \phi_2$ asserts that $\phi_2$ will be eventually satisfied and all preceding states satisfy $\phi_1$. The semantics of PCTL is defined by a satisfaction relation $\models$ as follows.

\vspace{-2mm}
\begin{eqnarray}
\centering
s&\models& true\ \ \text{iff state $s\in S$}\\
s&\models& \phi\qquad \text{iff state s satisfies the state formula $\phi$}\\
\tau&\models& \psi\qquad \text{iff trajectory $\tau$ satisfies the path formula $\psi$}.
\end{eqnarray} 
\vspace{-2mm}

Additionally, $\models_{min}$ denotes the minimal satisfaction relation~\cite{4770111} between $\tau$ and $\psi$. Defining $pref(\tau)$ as the set of all prefixes of trajectory $\tau$ including $\tau$ itself, then $\tau\models_{min} \psi$ iff $(\tau\models\psi) \wedge (\forall \tau'\in pref(\tau)\backslash\tau, \tau' \nvDash \psi)$. For instance, if $\psi=\phi_1 \until^{\leq k}\phi_2$, then for any finite trajectory $\tau\models_{min}\phi_1 \until^{\leq k}\phi_2$, only the final state in $\tau$ satisfies $\phi_2$. Let $P(\tau)$ be the probability of transitioning along a trajectory $\tau$ and let $\Gamma_\psi$ be the set of all finite trajectories that satisfy $\tau\models_{min}\psi$, the value of PCTL property $\psi$ is defined as $P_{=?|s_0}[\psi]=\sum\limits_{\tau\in\Gamma_\psi}P(\tau)$. For a DTMC $M_{\pi}$ and a state formula $\phi= P_{\leq p^*}[\psi]$, $M_{\pi} \models \phi$ iff $P_{=?|s_0}[\psi]\leq p^*$. 
  
A {\it counterexample} of $\phi$ is a set $cex\subseteq\Gamma_\psi$ that satisfies $\sum\limits_{\tau\in cex}P(\tau)> p^*$. 
Let $\mathbb{P}(\Gamma) = \sum\limits_{\tau\in \Gamma}P(\tau)$ be the sum of probabilities of all trajectories in a set $\Gamma$. 
Let $CEX_{\phi}\subseteq 2^{\Gamma_\psi}$ be the set of all counterexamples for a formula $\phi$ such that 
$(\forall cex\in CEX_{\phi},\mathbb{P}(cex)> p^*)$ and $(\forall \Gamma\in 2^{\Gamma_\psi}\backslash CEX_{\phi}, \mathbb{P}(\Gamma)\leq p^*)$. 
A \textit{minimal counterexample} is a set $cex\in CEX_{\phi}$ such that $\forall cex'\in CEX_{\phi}, |cex|\leq|cex'|$. 
By converting DTMC $M_\pi$ into a weighted directed graph, counterexample can be found by solving a k-shortest paths (KSP) problem or 
a hop-constrained KSP (HKSP) problem~\cite{4770111}. 
Alternatively, counterexamples can be found by using Satisfiability Modulo Theory solving or mixed integer linear programming to 
determine the minimal critical subsystems that capture the counterexamples in $M_\pi$~\cite{Wimmer2012}. 

A policy can also be synthesized by solving the objective $\underset{\pi}{min}\ P_{=?}[\psi]$ for an MDP $M$.
This problem can be solved by linear programming or policy iteration (and value iteration for step-bounded reachability)~\cite{Kwiatkowska2013}.

\section{Problem Formulation and Overview}
\label{sec:overview}
\graphicspath{{./}}
Suppose there are some unsafe states in an $MDP\backslash R$ $M = ($$S,\ $$ A,\ $$ T,\ $$ \gamma,\ $$s_0)$. 
A safety issue in apprenticeship learning means that an agent following the learnt policy would have a higher probability of entering those unsafe states than it should. There are multiple reasons that can give rise to this issue. First, it is possible that the expert policy $\pi_E$ itself has a high probability of reaching the unsafe states. Second, human experts often tend to perform only successful demonstrations that do not highlight the unwanted situations~\cite{shiarlis2016inverse}. This {\it lack of negative examples} in the training set can cause the learning agent to be unaware of the existence of those unsafe states.
\vspace{-3mm}
\begin{figure}[!htb]
\centering
\subfigure[]{
	\begin{minipage}[c][0.6\width]{
	   0.22\textwidth}
	   \includegraphics[width=1.\linewidth]{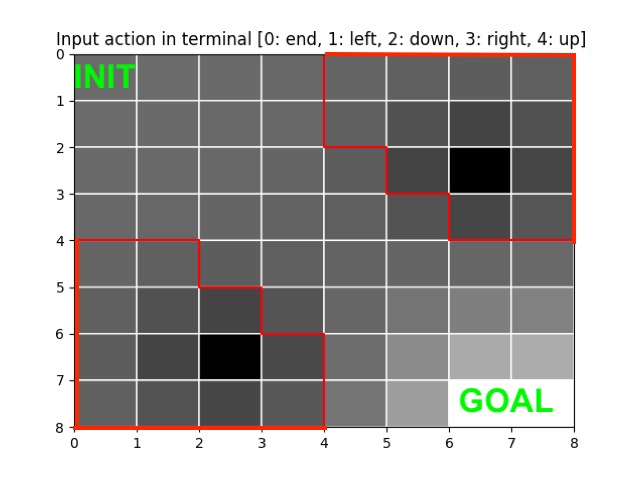}
	   \label{fig:sec3_2}
	\end{minipage}\hfill
}
\subfigure[]{
	\begin{minipage}[c][0.6\width]{
	   0.22\textwidth}
	   \includegraphics[width=1.\linewidth]{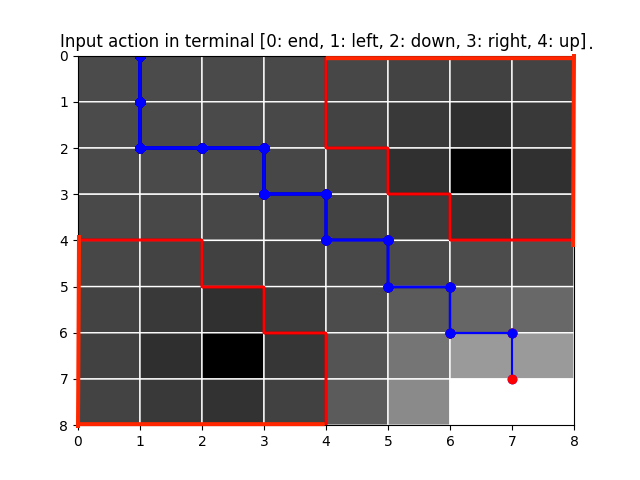}
	   \label{fig:sec3_3}
	\end{minipage}\hfill
}
\subfigure[]{
	\begin{minipage}[c][0.6\width]{
	   0.22\textwidth}
	   \includegraphics[width=1.\linewidth]{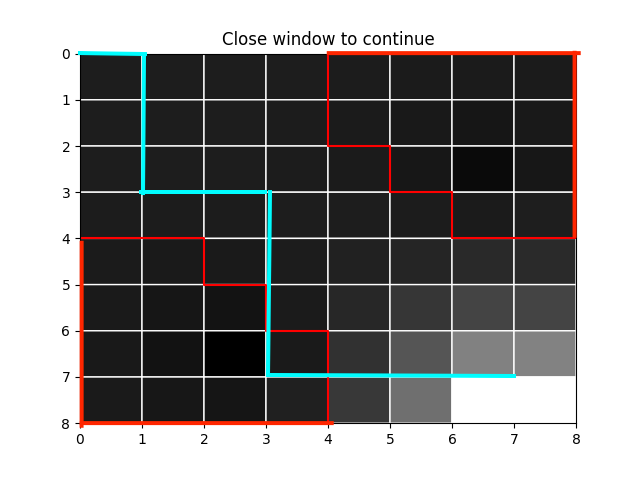}
	   \label{fig:sec3_5}
	\end{minipage}\hfill
}
\subfigure[]{
	\begin{minipage}[c][0.6\width]{
	   0.22\textwidth}
	   \includegraphics[width=1.\linewidth]{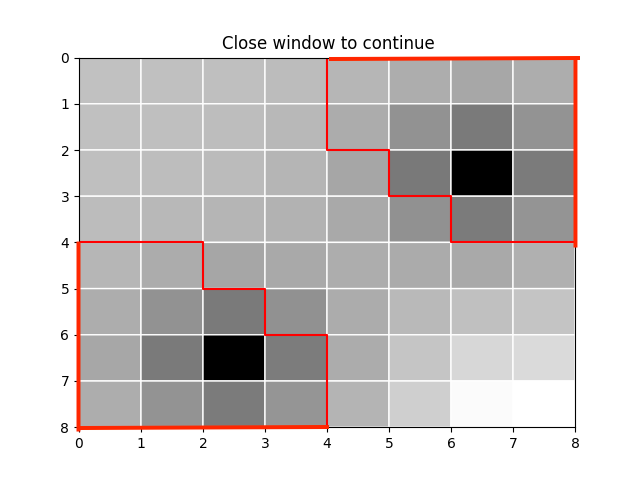}
	   \label{fig:sec3_4}
	\end{minipage}\hspace{0.012\linewidth}
}
\vspace{-2mm}
\caption{
The 8 x 8 grid-world. (a) Lighter grid cells have higher rewards than the darker ones. The two black grid cells have the lowest rewards, while the two white ones have the highest rewards. The grid cells enclosed by red lines are considered \textit{unsafe}. (b) The blue line is an example trajectory demonstrated by the expert. (c) Only the goal states are assigned high rewards and there is little difference between the unsafe states and their nearby states. As a result, the learnt policy has a high probability of passing through the unsafe states as indicated by the cyan line. (d) $p^*=20\%$. The learnt policy is optimal to a reward function that correctly assigns low rewards to the unsafe states. 
}
\label{fig:grid_world1}
\end{figure}
\vspace{-5mm}

We use a 8 x 8 grid-world navigation example as shown in Fig.~\ref{fig:grid_world1} to illustrate this problem. An agent starts from the upper-left corner and moves from cell to cell until it reaches the lower-right corner. The `unsafe' cells are enclosed by the red lines. These represent regions that the agent should avoid. In each step, the agent can choose to stay in current cell or move to an adjacent cell but with $20\%$ chance of moving randomly instead of following its decision. The goal area, the unsafe area and the reward mapping for all states are shown in Fig.~\ref{fig:sec3_2}. For each state $s\in S$, its feature vector consists of $4$ radial basis feature functions with respect to the squared Euclidean distances between $s$ and the $4$ states with the highest or lowest rewards as shown in Fig.~\ref{fig:sec3_2}. 
In addition, a specification $\Phi$ formalized in PCTL is used  to capture the safety requirement. In (\ref{eq:example-spec}), $p^*$ is the required upper bound of the probability of reaching an unsafe state within $t=64$ steps. 
\begin{equation}
\Phi ::= P_{\leq p^*}[\true\ \until^{\leq t}\ \texttt{unsafe}]
\label{eq:example-spec}
\end{equation}

Let $\pi_E$ be the optimal policy under the reward map shown in Fig.~\ref{fig:sec3_2}. 
The probability of entering an unsafe region within $64$ steps by following $\pi_E$ is $24.6\%$. 
Now consider the scenario where the expert performs a number of demonstrations by following $\pi_E$. 
{\it All demonstrated trajectories in this case successfully reach the goal areas without ever passing through any of the  unsafe regions.}
Fig.~\ref{fig:sec3_3} shows a representative trajectory (in blue) among $10,000$ such demonstrated trajectories.
The resulting reward map by running the AL algorithm on these $10,000$ demonstrations is shown in Fig.~{\ref{fig:sec3_5}}.
Observe that only the goal area has been learnt whereas the agent is oblivious to the unsafe regions (treating them in the same way as other dark cells). 
In fact, the probability of reaching an unsafe state within $64$ steps with this policy turns out to be $82.6\%$ (thus violating the safety requirement by a large margin).  
To make matters worse, the value of $p^*$ may be decided or revised after a policy has been learnt. 
In those cases, even the original expert policy $\pi_E$ may be unsafe, e.g., when $p^*=20\%$. 
Thus, we need to adapt the original AL algorithm so that it will take into account of such safety requirement.
Fig.~\ref{fig:sec3_4} shows the resulting reward map learned using our proposed algorithm (to be described in detail later) for $p^*=20\%$. 
It clearly matches well with the color differentiation in the original reward map and captures both the goal states and the unsafe regions. This policy has an unsafe probability of $19.0\%$.
We are now ready to state our problem. 

\begin{definition}
\textbf{The safety-aware apprenticeship learning (SafeAL) problem} is, given an $MDP\backslash R$, a set of $m$ trajectories $\{\tau_0, \tau_1, ..., \tau_{m-1}\}$ demonstrated by an expert, and a specification $\Phi$, to learn a policy $\pi$ that satisfies $\Phi$ and is $\epsilon$-close to the expert policy $\pi_E$.
\end{definition}

\begin{remark}
We note that a solution may not always exist for the SafeAL problem. While the decision problem of checking whether a solution exists is of theoretical interest, 
in this paper, we focus on tackling the problem of finding a policy $\pi$ that satisfies a PCTL formula $\Phi$ (if $\Phi$ is satisfiable) and 
whose performance is as close to that of the expert's as possible, i.e. we relax the condition on $\mu_{\pi}$ being $\epsilon$-close to $\mu_E$. 
\end{remark}

\section{A Framework for Safety-Aware Learning}
\label{sec:counterexample}
\begin{figure}
\centering
  \includegraphics[width=7cm, height=7cm]{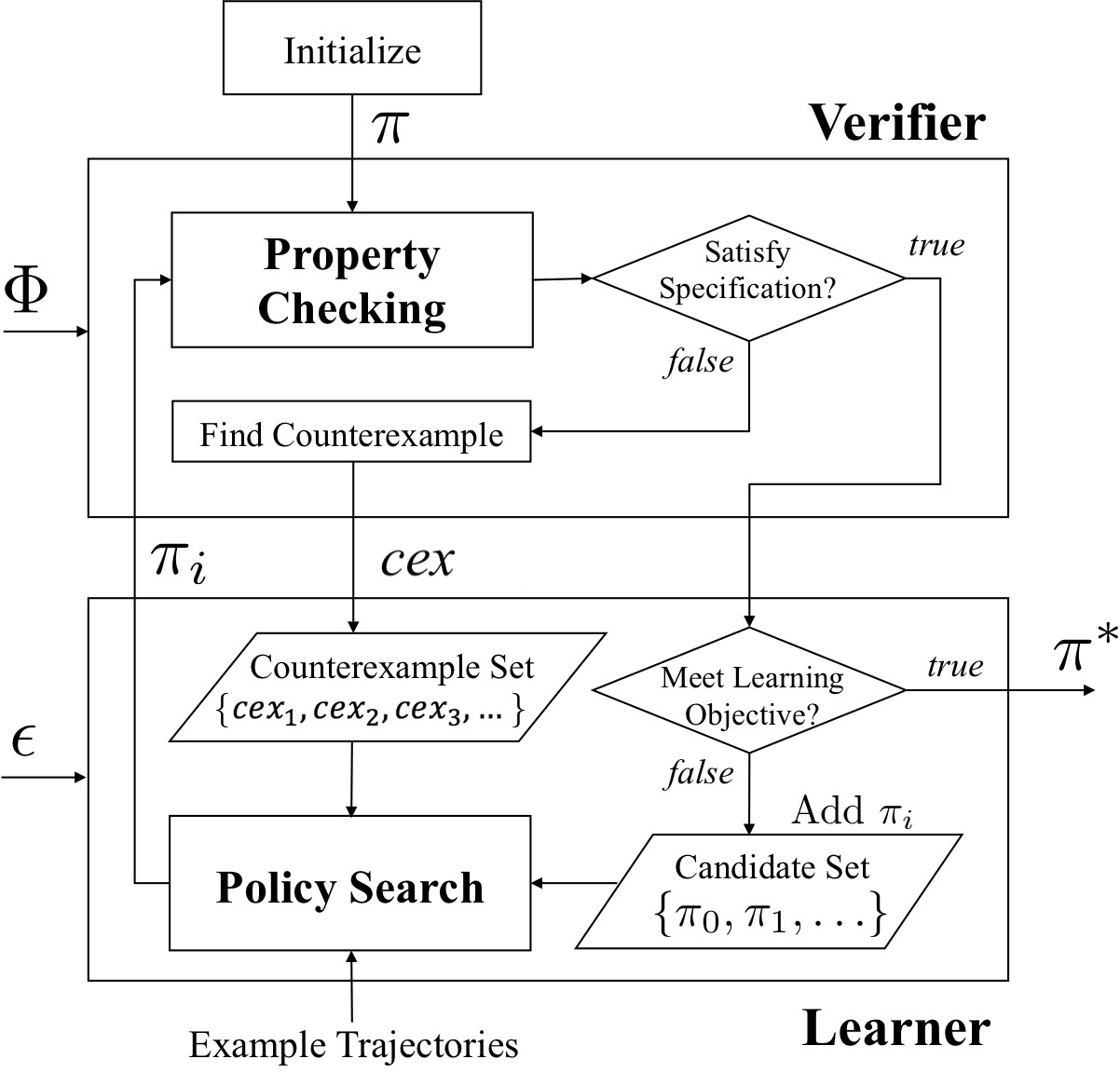}
  \vspace{-2mm}
\caption{
Our safety-aware learning framework. Given an initial policy $\pi_0$, a specification $\Phi$ and a learning objective (as captured by $\epsilon$), the framework iterates between a {\it verifier} and a {\it learner} to search for a policy $\pi^*$ that satisfies both $\Phi$ and $\epsilon$. One invariant that this framework maintains is that all the $\pi_i$'s in the candidate policy set satisfy $\Phi$.}
\label{fig:sec4_1}
\end{figure}

\graphicspath{{./}}
In this section, we describe a general framework for safety-aware learning. 
This novel framework utilizes information from both the expert demonstrations and a verifier. 
The proposed framework is illustrated in Fig.~\ref{fig:sec4_1}. Similar to the \emph{counterexample-guided inductive synthesis} (CEGIS) paradigm~\cite{CEGIS}, our framework consists of a {\it verifier} and a {\it learner}. The verifier checks if a candidate policy satisfies the safety specification $\Phi$. In case $\Phi$ is not satisfied, the verifier generates a counterexample for $\Phi$. 
The main difference from CEGIS is that our framework considers not only functional correctness, e.g., safety, but also performance (as captured by the learning objective). Starting from an initial policy $\pi_0$, each time the learner learns a new policy, the verifier checks if the specification is satisfied. If true, then this policy is added to the candidate set, otherwise the verifier will generate a (minimal) counterexample and add it to the counterexample set. During the learning phase, the learner uses both the counterexample set and candidate set to find a policy that is close to the (unknown) expert policy and far away from the counterexamples. The goal is to find a policy that is $\epsilon$-close to the expert policy and satisfies the specification. 
For the grid-world example introduced in Section~\ref{sec:overview}, when $p^*=5\%$ (thus presenting a stricter safety requirement compared to the expert policy $\pi_E$), our approach produces a policy with only $4.2\%$ of reaching an unsafe state within $64$ steps (with the correspondingly inferred reward mapping shown in Fig.~\ref{fig:sec3_4}). 
\vspace{-3mm}
\vspace{-7mm}
\begin{figure}[h]
\centering
\subfigure[]{
	\hspace{0.1\linewidth}\begin{minipage}[c][0.8\width]{
	   0.25\textwidth}
	   \includegraphics[width=1.3\linewidth]{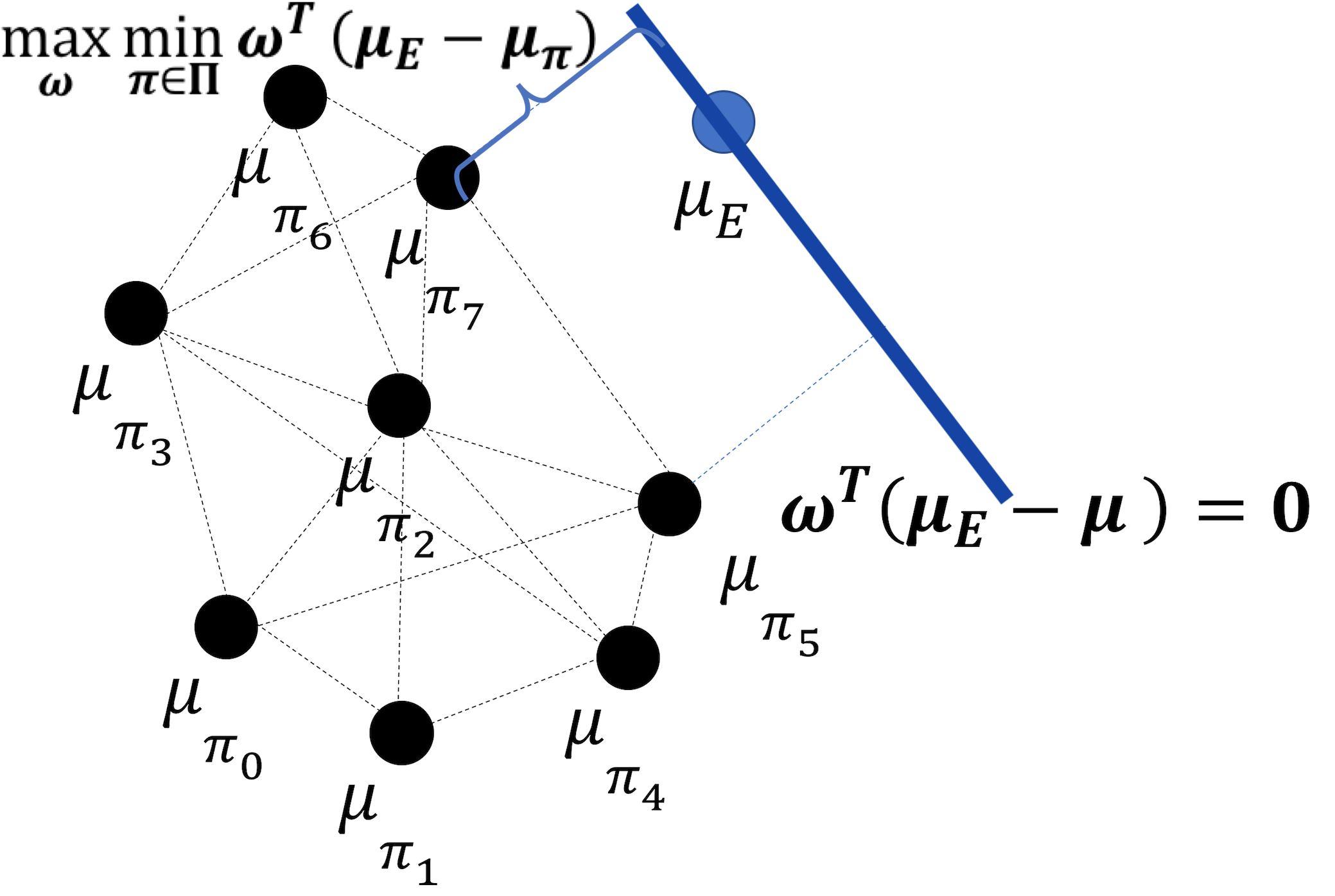}
	   \label{fig:sec4_2}
	\end{minipage}\hspace{0.2\linewidth}
}
\subfigure[]{
	\begin{minipage}[c][0.8\width]{
	   0.25\textwidth}
	   \includegraphics[width=1.3\linewidth]{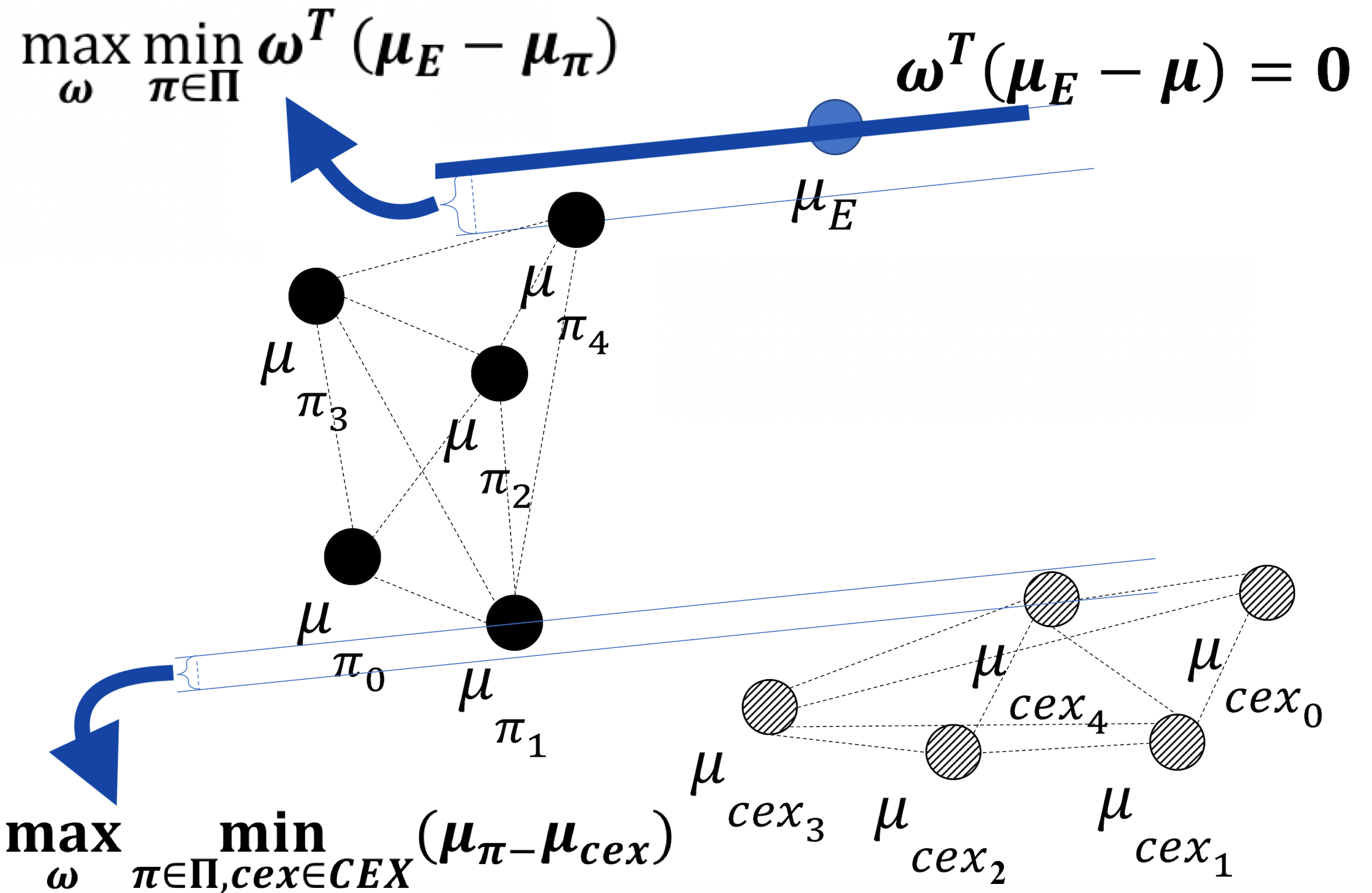}
	   \label{fig:sec4_3}
	\end{minipage}\hspace{0.1\linewidth}
}
\vspace{-3mm} 
\caption{(a) Learn from expert. (b) Learn from both expert demonstrations and counterexamples.}
\vspace{-4mm}
\end{figure}

Learning from a (minimal) counterexample $cex_{\pi}$ of a policy $\pi$ is similar to learning from expert demonstrations. 
The basic principle of the AL algorithm proposed in \cite{Abbeel:2004:ALV:1015330.1015430} is to find a weight vector $\omega$ under which the expected reward of $\pi_E$ maximally outperforms any mixture of the policies in the candidate policy set $\Pi=\{\pi_0, \pi_1, \pi_2, \ldots\}$. Thus, $\omega$ can be viewed as the normal vector of the hyperplane $\omega^T(\mu - \mu_E) = 0$ that has the maximal distance to the convex hull of the set $\{\mu_{\pi}\:|\:\pi\in\Pi\}$ as illustrated in the 2D feature space in Fig.~\ref{fig:sec4_2}. 
It can be shown that $\omega^T \mu_\pi \geq \omega^T \mu_{\pi'}$ for all previously found $\pi'$s. Intuitively, this helps to move the candidate $\mu_\pi$ closer to $\mu_E$.
Similarly, we can apply the same max-margin separation principle to maximize the distance between the candidate policies and the counterexamples (in the $\mu$ space).     
Let ${CEX}= \{cex_0, cex_1, cex_2, ...\}$ denote the set of counterexamples of the policies that do not satisfy the specification $\Phi$. 
Maximizing the distance between the convex hulls of the sets $\{\mu_{cex}\:|\:cex\in{CEX}\}$ and $\{\mu_{\pi}\:|\:\pi\in\Pi\}$ is equivalent to maximizing the distance between the parallel supporting hyperplanes of the two convex hulls as shown in Fig.~\ref{fig:sec4_3}. The corresponding optimization function is given in Eq.~(\ref{eq:sec4_2}).

\vspace{-2mm}
\begin{equation}
\delta = \max\limits_{\omega}\min\limits_{\pi\in\Pi, cex\in CEX}\ \omega^T(\mu_{\pi}-\mu_{cex})\label{eq:sec4_2}\qquad s.t.\:||\omega||_2\leq 1
\end{equation}
\vspace{-2mm}

To attain good performance similar to that of the expert, we still want to learn from $\mu_E$. Thus, the overall problem can be formulated as a multi-objective optimization problem that combines (\ref{eq:sec1_1}) and (\ref{eq:sec4_2}) into (\ref{eq:sec4_3}).

\vspace{-3mm}
\begin{equation}
\max\limits_\omega \min\limits_{\pi \in \Pi, \tilde\pi \in \Pi, cex \in CEX} (\omega^T (\mu_E - \mu_{\pi}),\ \omega^T (\mu_{\tilde\pi} - \mu_{cex}) )\qquad s.t.\:||\omega||_2\leq 1
\label{eq:sec4_3}
\end{equation}
\vspace{-5mm}

\section{Counterexample-Guided Apprenticeship Learning}
\label{sec:safeAL}
In this section, we introduce the CounterExample Guided Apprenticeship Learning (CEGAL)
algorithm to solve the SafeAL problem. 
It can be viewed as a special case of the safety-aware learning framework described in the previous section. 
In addition to combining policy verification, counterexample generation and AL, our approach uses an adaptive weighting scheme to weight the separation from $\mu_E$ with the separation from $\mu_{cex}$.

\vspace{-4mm}
\begin{eqnarray}
&&\underset{\omega}{\max}\underset{\pi\in\Pi_S,\tilde{\pi}\in\Pi_S, cex\in CEX}{\min}\ \omega^T(k(\mu_E - \mu_{\pi})+(1-k)(\mu_{ \tilde{\pi}}  - \mu_{cex}))\label{eq:sec5_1}\\
&&s.t.\: ||\omega||_2\leq 1 \label{eq:sec5_4},\: k\in[0, 1]\label{eq:sec5_5} \nonumber\\ 
&&\quad\ \:\omega^T(\mu_E - \mu_{\pi})\leq\omega^T(\mu_E - \mu_{\pi'}),\ \forall\pi'\in\Pi_S \label{eq:sec5_2} \nonumber\\ 
&&\quad\ \:\omega^T(\mu_{\tilde{\pi}} - \mu_{cex})\leq\omega^T(\mu_{ \tilde{\pi}'} - \mu_{cex'}),\ \forall \tilde{\pi}'\in\Pi_S, \forall cex'\in{CEX} \nonumber
\end{eqnarray}
\vspace{-3mm}

In essence, we take a weighted-sum approach for solving the multi-objective optimization problem  (\ref{eq:sec4_3}). Assuming that $\Pi_S=\{\pi_{1},$ $\pi_{2},$ $\pi_{3},$ $\ldots \}$ is a set of candidate policies that all satisfy $\Phi$, ${CEX} =\{cex_1,$ $cex_2,$ $cex_3,$ $\ldots\}$ is a set of counterexamples. We introduce a parameter $k$ and change (\ref{eq:sec4_3}) into a weighted sum optimization problem (\ref{eq:sec5_1}). Note that $\pi$ and $\tilde\pi$ can be different. The optimal $\omega$ solved from (\ref{eq:sec5_1}) can be used to generate a new policy $\pi_\omega$ by using algorithms such as policy iteration. 
We use a probabilistic model checker, such as PRISM~\cite{kwiatkowska2002prism}, to check if $\pi_\omega$ satisfies $\Phi$. If it does, then it will be added to $\Pi_S$. Otherwise, a counterexample generator, such as COMICS~\cite{DBLP:journals/corr/abs-1206-0603}, is used to generate a (minimal) counterexample $cex_{\pi_\omega}$, which will be added to ${CEX}$.  

\vspace{-2mm}
\begin{algorithm}[htb]
\caption{Counterexample-Guided Apprenticeship Learning (CEGAL)}
\begin{algorithmic}[1]
\State \textbf{Input}:
\\\qquad $M\gets$ A partially known $MDP\backslash R$; $f\gets$ A vector of feature functions
\\\qquad $\mu_E\gets$ The expected features of expert trajectories $\{\tau_0, \tau_1, \ldots,\tau_m\}$
\\\qquad $\Phi\gets$ Specification; $\epsilon\gets$ Error bound for the expected features;
\\\qquad $\sigma, \alpha\in(0, 1)\gets$ Error bound $\sigma$ and step length $\alpha$ for the parameter $k$; 
\State \textbf{Initialization}:
\\\qquad \textbf{If} $||\mu_E-\mu_{\pi_0}||_2\leq \epsilon$, {\bf then return} $\pi_0$ \Comment{$\pi_0$ is the \textbf{initial safe policy}}
\\\qquad $\Pi_S\leftarrow\{\pi_0\}$, $CEX\leftarrow\emptyset$ \Comment{Initialize candidate and counterexample set}
\\\qquad $inf\leftarrow0, sup\leftarrow1, k\leftarrow sup$ \Comment{Initialize multi-optimization parameter $k$}
\\\qquad $\pi_1\leftarrow$ Policy learnt from $\mu_E$ via apprenticeship learning
\State \textbf{Iteration $i\ (i\geq 1)$}:
\\\qquad\textbf{Verifier:}
\\\qquad\qquad status $\gets Model\_Checker(M,\pi_i, \Phi)$
\\\qquad\qquad \textbf{If} status = SAT, \textbf{then go to Learner}
\\\qquad\qquad \textbf{If} status = UNSAT
\\\qquad\qquad\qquad $cex_{\pi_i}\gets Counterexample\_Generator(M,\pi_i,\Phi)$
\\\qquad\qquad\qquad Add $cex_{\pi_i}$ to $CEX$ and solve $\mu_{cex_{\pi_i}}$, \textbf{go to Learner} 
\\\qquad\textbf{Learner:}
\\\qquad\qquad \textbf{If} status = SAT
\\\qquad\qquad\qquad \textbf{If} $||\mu_E-\mu_{\pi_i}||_2\leq \epsilon$, {\bf then return} $\pi^*\gets\pi_i$
\\\Comment{Terminate. $\pi_i$ is {$\epsilon$-close} to $\pi_E$}
\\\qquad\qquad\qquad Add $\pi_i$ to $\Pi_S$, $inf \leftarrow k, k \leftarrow sup$\Comment{Update $\Pi_S$, $inf$ and reset $k$}
\\\qquad\qquad \textbf{If} status = UNSAT
\\\qquad\qquad\qquad \textbf{If} $|k - inf|\leq\sigma$, {\bf then return} $\pi^*\leftarrow\underset{{\pi}\in \Pi_S}{argmin}||\mu_E - \mu_{{\pi}}||_2$\\\Comment{Terminate. $k$ is too close to its lower bound. }
\\\qquad\qquad\qquad $k \leftarrow \alpha\cdot inf + (1 - \alpha)k$ 
\Comment{Decrease k to learn for safety}
\\\qquad\qquad $\omega_{i+1} \leftarrow arg\underset{\omega}{max}\underset{\pi\in\Pi_S, \tilde{\pi}\in\Pi_S, cex\in CEX}{\min}\ \omega^T(k(\mu_E - \mu_{\pi})+(1-k)(\mu_{\tilde{\pi}}  - \mu_{cex}))$
\\\Comment{Note that the multi-objective optimization function recovers AL when $k=1$} 
\\\qquad\qquad $\pi_{i+1}, \mu_{\pi_{i+1}}\gets$ Compute the optimal policy $\pi_{i+1}$ and its expected features $\mu_{\pi_{i+1}}$ for the MDP $M$ with reward $R(s)=\omega_{i+1}^T f(s)$
\\\qquad{\bf Go to next iteration}
\end{algorithmic}
\label{algo1}
\end{algorithm}

Algorithm 1 describes CEGAL in detail. 
With a constant $sup=1$ and a variable $inf\in[0, sup]$ for the upper and lower bounds respectively, the learner determines the value of $k$ within $[inf, sup]$ in each iteration depending on the outcome of the verifier and uses $k$ in solving (\ref{eq:sec5_1}) in line 27. 
Like most nonlinear optimization algorithms, this algorithm requires an initial guess, which is an initial safe policy $\pi_0$ to make $\Pi_S$ nonempty. A good initial candidate would be the maximally safe policy for example obtained using PRISM-games~{\cite{Kwiatkowska2017}}. 
Without loss of generality, we assume this policy satisfies $\Phi$.
Suppose in iteration $i$, an intermediate policy $\pi_i$ learnt by the learner in iteration $i-1$ is verified to satisfy $\Phi$, then we increase $inf$ to $inf=k$ and reset $k$ to $k=sup$ as shown in line 22. 
If $\pi_i$ does not satisfy $\Phi$, then we reduce $k$ to $k=\alpha\cdot inf + (1 - \alpha)k$  as shown in line 26 where $\alpha\in(0, 1)$ is a step length parameter. 
If $|k-inf|\leq\sigma$ and $\pi_i$ still does not satisfy $\Phi$, the algorithm chooses from $\Pi_S$ a best safe policy $\pi^*$ which has the smallest margin to $\pi_E$ as shown in line 24. If $\pi_i$ satisfies $\Phi$ and is {$\epsilon$-close} to $\pi_E$, the algorithm outputs $\pi_i$ as show in line 19. For the occasions when $\pi_i$ satisfies $\Phi$ and $inf = sup = k = 1$, solving (\ref{eq:sec5_1}) is equivalent to solving (\ref{eq:sec1_1}) as in the original AL algorithm.

\begin{remark}
The initial policy $\pi_0$ does not have to be maximally safe, although such a policy can be used to verify if $\Phi$ is  satisfiable at all. 
Naively safe policies often suffice for obtaining a safe and performant output at the end. Such a policy can be obtained easily in many settings, e.g., in the grid-world example one safe policy is simply staying in the initial cell. 
In both cases, $\pi_0$ typically has very low performance since satisfying $\Phi$ is the only requirement for it. 
\end{remark}

\begin{theorem}
Given an initial policy $\pi_0$ that satisfies $\Phi$, Algorithm~\ref{algo1} is guaranteed to output a policy $\pi^*$, such that (1) $\pi^*$ satisfies $\Phi$, and 
(2) the performance of $\pi^*$ is at least as good as that of $\pi_0$ when compared to $\pi_E$, i.e. $\|\mu_E - \mu_{\pi^*}\|_2\leq\|\mu_E - \mu_{\pi_0}\|_2$. 
\end{theorem}
\noindent{\it Proof sketch. }
The first part of the guarantee can be proven by case splitting. 
Algorithm~\ref{algo1} outputs $\pi^*$ either when $\pi^*$ satisfies $\Phi$ and is {$\epsilon$-close} to $\pi_E$, or when $|k-inf|\leq \sigma$ in some iteration. 
In the first case, $\pi^*$ clearly satisfies $\Phi$. 
In the second case, $\pi^*$ is selected from the set $\Pi_S$ which contains all the policies that have been found to satisfy $\Phi$ so far, so $\pi^*$ satisfies $\Phi$.
For the second part of the guarantee, the initial policy $\pi_0$ is the final output $\pi^*$ if $\pi_0$ satisfies $\Phi$ and is {$\epsilon$-close} to $\pi_E$. 
Otherwise, $\pi_0$ is added to $\Pi_S$ if it satisfies $\Phi$.  
During the iteration, if $|k-inf|\leq \sigma$ in some iteration, then the final output is $\pi^*=\underset{{\pi}\in\Pi_S}{argmin}||\mu_E - \mu_{{\pi}}||_2$, so it must satisfy $\|\mu_E -  \mu_{\pi^*}\|_2\leq\|\mu_E - \mu_{\pi_0}\|_2$. 
If a learnt policy $\pi^*$ satisfies $\Phi$ and is {$\epsilon$-close} to $\pi_E$, then Algorithm~\ref{algo1} outputs $\pi^*$ without adding it to $\Pi_S$. Obviously $\|\mu_E - \mu_{{\pi}}\|_2>\epsilon, \forall{\pi}\in\Pi_S$, so $\|\mu_E -  \mu_{\pi^*}\|_2\leq\|\mu_E - \mu_{\pi_0}\|_2$.

\noindent
{\it Discussion.} In the worst case, CEGAL will return the initial safe policy. However, this can be because a policy that simultaneously satisfies $\Phi$ and is $\epsilon$-close to the expert's demonstrations does not exist. Comparing to AL which offers no safety guarantee and finding the maximally safe policy which has very poor performance, CEGAL provides a principled way of guaranteeing safety while retaining performance.

\noindent
{\it Convergence.} 
Algorithm 1 is guaranteed to terminate. Let $inf_t$ be the $t^{th}$ assigned value of $inf$. After $inf_t$ is given, $k$ is decreased from $k_0=sup$ iteratively by $k_{i}=\alpha\cdot inf_t + (1 - \alpha)k_{i-1}$ until either $|k_i-inf_t|\leq \sigma$ in line 24 or a new safe policy is found in line 18. The update of $k$ satisfies the following equality.

\vspace{-3mm}
\begin{eqnarray}
\frac{|k_{i+1} - inf_t|}{|k_i - inf_t|}&=& \frac{\alpha\cdot inf_t + (1 - \alpha)k_i - inf_t}{k_i - inf_t} =  1-\alpha
\end{eqnarray}
\vspace{-3mm}

Thus, it takes no more than 1+$\log_{1-\alpha}\frac{\sigma}{sup - inf_t}$ iterations for either the algorithm to terminate in line 24 or a new safe policy to be found in line 18. If a new safe policy is found in line 18, $inf$ will be assigned in line 22 by the current value of $k$ as $inf_{t+1}=k$ which obviously satisfies $inf_{t+1} - inf_t \geq(1-\alpha)\sigma$. After the assignment of $inf_{t+1}$, the iterative update of $k$ resumes. Since $sup-inf_t \leq 1$, the following inequality holds.

\vspace{-3mm}
\begin{eqnarray}
\frac{|inf_{t+1} - sup|}{|inf_{t} - sup|}&\leq&\frac{sup -inf_{t} - (1-\alpha)\sigma}{sup- inf_{t}}\leq 1 - (1-\alpha)\sigma
\end{eqnarray}
\vspace{-3mm}

Obviously, starting from an initial $inf=inf_0<sup$, with the alternating update of $inf$ and $k$, $inf$ will keep getting close to $sup$ unless the algorithm terminates as in line 24 or a safe policy {$\epsilon$-close} to $\pi_E$ is found as in line 19. The extreme case is that finally $inf=sup$ after no more than $\frac{sup-inf_0}{(1-\alpha)\sigma}$ updates on $inf$. Then, the problem becomes AL. Therefore, the worst case of this algorithm can have two phases. In the first phase, $inf$ increases from $inf=0$ to $inf=sup$. Between each two consecutive updates $(t, t+1)$ on $inf$, there are no more than $\log_{1-\alpha}\frac{(1-\alpha)\sigma}{sup - inf_t}$ updates on $k$ before $inf$ is increased from $inf_t$ to $inf_{t+1}$. Overall, this phase takes no more than

\vspace{-3mm}
\begin{equation}
{\underset{0\leq i< \frac{sup-inf_0}{(1-\alpha)\sigma}}{\sum}\log_{1-\alpha} \frac{(1-\alpha)\sigma}{sup-inf_0-i\cdot (1-\alpha)\sigma}} = {\underset{0\leq i< \frac{1}{(1-\alpha)\sigma}}{\sum}\log_{1-\alpha} \frac{(1-\alpha)\sigma}{1-i\cdot (1-\alpha)\sigma}}
\end{equation}
\vspace{-3mm}

\noindent iterations to reduce the multi-objective optimization problem to original apprenticeship learning and then the second phase begins. Since $k=sup$, the iteration will stop immediately when an unsafe policy is learnt as in line 24. This phase will not take more iterations than original AL algorithm does to converge and the convergence result of AL is given in \cite{Abbeel:2004:ALV:1015330.1015430}. 

In each iteration, the algorithm first solves a second-order cone programming (SOCP) problem (\ref{eq:sec5_1}) to learn a policy. SOCP problems can be solved in polynomial time by interior-point (IP) methods~\cite{Kuo:2004aa}.
PCTL model checking for DTMCs can be solved in time linear in the size of the formula and polynomial in the size of the state space~\cite{Hansson1994}. 
Counterexample generation can be done either by enumerating paths using the $k$-shortest path algorithm or determining a critical subsystem using either a $SMT$ formulation or mixed integer linear programming (MILP)~\cite{Wimmer2012}. For the $k$-shortest path-based algorithm, it can be computationally expensive sometimes to enumerate a large amount of paths (i.e. a large $k$) when $p^*$ is large. This can be alleviated by using a smaller $p^*$ during calculation, which is equivalent to considering only paths that have high probabilities.
\vspace{-1mm}

\section{Experiments}
\label{sec:exp}
\graphicspath{{./}}
\vspace{-1mm}
We evaluate our algorithm on three case studies: (1) grid-world, (2) cart-pole, and (3) mountain-car. The cart-pole environment\footnote{\url{https://github.com/openai/gym/wiki/CartPole-v0}} and the mountain-car environment\footnote{\url{https://github.com/openai/gym/wiki/MountainCar-v0}} are obtained from OpenAI Gym. All experiments are carried out on a quad-core i7-7700K processor running at 3.6 GHz with 16 GB of memory. 
Our prototype tool was implemented in Python\footnote{\url{https://github.com/zwc662/CAV2018}}.
The parameters are $\gamma=0.99, \epsilon=10, \sigma=10^{-5}, \alpha=0.5$ and the maximum number of iterations is $50$.
For the OpenAI-gym experiments, in each step, the agent sends an action to the OpenAI environment and the environment returns an observation and a reward (0 or 1). 
We show that our algorithm can guarantee safety while retaining the performance of the learnt policy compared with using AL alone. 
\vspace{-2mm}
\subsection{Grid World}
We first evaluate the scalability of our tool using the grid-world example.
Table~\ref{tab:grid-world} shows the average runtime (per iteration) for the individual components of our tool as the size of the grid-world increases. The first and second columns indicate the size of the grid world and the resulting state space. The third column shows the average runtime that policy iteration takes to compute an optimal policy $\pi$ for a known reward function. The forth column indicates the average runtime that policy iteration takes to compute the expected features $\mu$ for a known policy. The fifth column indicates the average runtime of verifying the PCTL formula using PRISM. The last column indicates the average runtime that generating a counterexample using COMICS.  
\vspace{-7mm}
\begin{table}[htb]
\begin{center}
\caption{Average runtime per iteration in seconds.}
\begin{tabular}{|c|r|r|r|r|r|}
\hline
Size & Num. of States & Compute $\pi$ & Compute $\mu$ & MC & Cex\\
\hline
$8 \times 8$ & 64 & 0.02 & 0.02 & 1.39 & 0.014\\
$16 \times 16$ & 256 & 0.05 & 0.05 & 1.43  & 0.014\\
$32 \times 32$ & 1024 & 0.07 & 0.08 & 3.12 & 0.035\\
$64 \times 64$ & 4096 & 6.52 & 25.88 &  22.877  & 1.59\\
\hline
\end{tabular}
\label{tab:grid-world}
\end{center}
\end{table}
\vspace{-10mm}

\subsection{Cart-Pole from OpenAI Gym}
\begin{figure}[hbt]
  \centering
   	\subfigure[]{
	\begin{minipage}[c][0.6\width]{
	   0.3\textwidth}
	   \includegraphics[width=\linewidth]{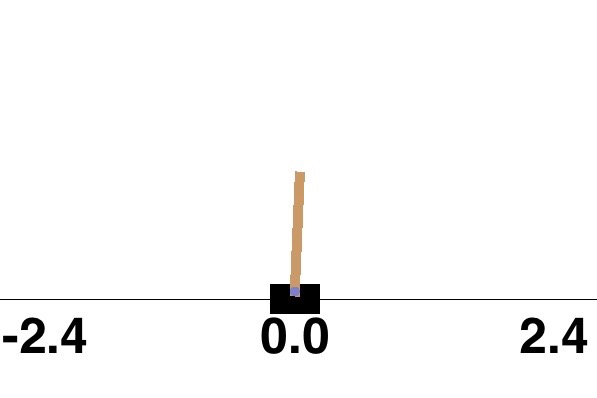}
	   \label{fig:cartpole-v0}
	\end{minipage}\hfill
}
\subfigure[]{
	\begin{minipage}[c][0.6\width]{
	   0.3\textwidth}
	   \includegraphics[width=\linewidth]{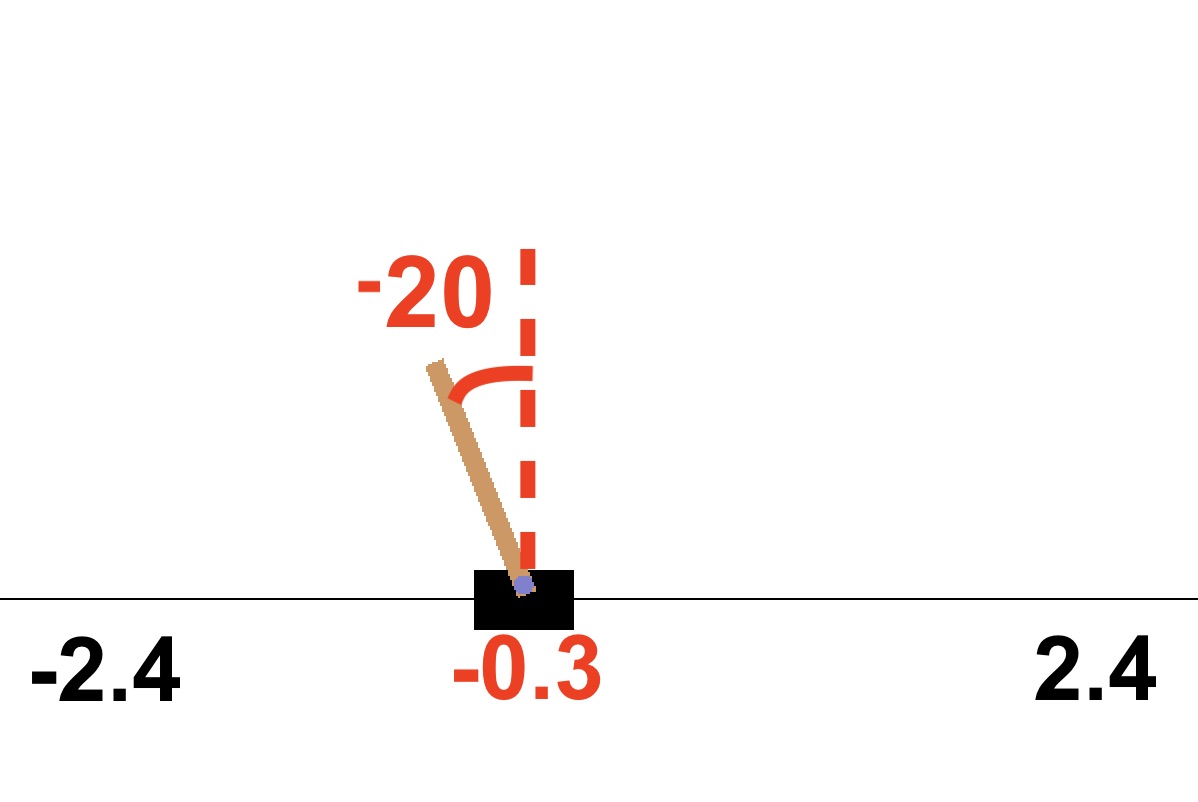}
	   \label{fig:cartpole-v1}
	\end{minipage}\hfill
}
\subfigure[]{
	\begin{minipage}[c][0.6\width]{
	   0.3\textwidth}
	   \includegraphics[width=\linewidth]{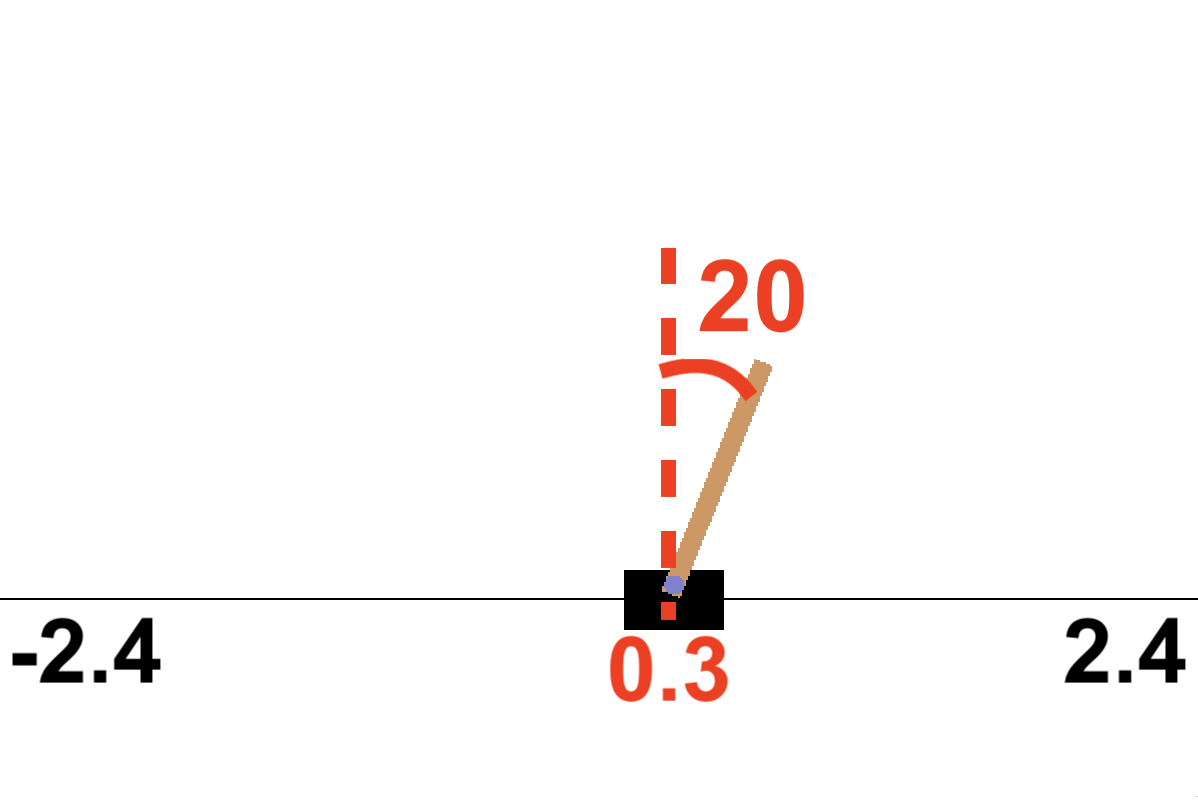}
	   \label{fig:cartpole-v2}
	\end{minipage}\hfill
}
\vspace{-3mm} 
  \caption{(a) The cart-pole environment. (b) The cart is at -0.3 and pole angle is \ang{-20}. (c) The cart is at 0.3 and pole angle is \ang{20}.}    
\label{fig:cartpole}
\vspace{-3mm}
\end{figure}

In the cart-pole environment as shown in Fig.~\ref{fig:cartpole-v0}, the goal is to keep the pole on a cart from falling over as long as possible by moving the cart either to the left or to the right in each time step. The maximum step length is $t=200$. The position, velocity and angle of the cart and the pole are continuous values and observable, but the actual dynamics of the system are unknown\footnote{The MDP is built from sampled data. The feature vector in each state contains $30$ radial basis functions which depend on the squared Euclidean distances between current state and other $30$ states which are uniformly distributed in the state space.}. 

A maneuver is deemed {\it unsafe} if the pole angle is larger than $\pm \ang{20}$ while the cart's horizontal position is more than $\pm 0.3$ as shown in Fig.~\ref{fig:cartpole-v1} and \ref{fig:cartpole-v2}. We formalize the safety requirement in PCTL as (\ref{eq:spec1}).
\vspace{-2mm}
\begin{equation}
\begin{split}
\Phi ::= P_{\leq p^*}[true\ \until^{\leq t}\ &(angle\leq \ang{-20}\wedge position\leq-0.3)\\
&\vee(angle\geq \ang{20}\wedge position\geq 0.3)]
\end{split}
\label{eq:spec1}
\end{equation}
\vspace{-12mm}

\begin{table}[htb]
\begin{center}
\caption{In the cart-pole environment, {\it higher} average steps mean better performance. The safest policy is synthesized using PRISM-games.}
\begin{tabular}{|r|r|r|r|r|}
\hline
\commentout{
  &\small{\ \ MC Result} &\small{\ \ Avg. Steps}  &\small{\ \ Unsafe Rate} &\small{\ \ Num. of Iters}\\
\hline
AL &49.1\%& 165 &  19\% & 2\\
\hline
Safest Policy  & 0.0\% & 8 & 0.0\% & N.A.\\
\hline
$p^*=30\%$ & 17.2\%  & 121 & 13.0\% & 10\\
\hline
$p^*=25\%$ & 9.3\%  & 136 & 17.0\% & 14\\
\hline
$p^*=20\%$ & 17.2\% & 122 & 10.8\% & 10\\
\hline
$p^*=15\%$ & 6.1\% & 120 & 5.4\% & 22\\
\hline
$p^*=10\%$ & 7.2\% & 136 & 13.7\% & 21\\
\hline
$p^*=5\%$ & 0.04\% & 83 & 0.5\% & 50\\
}
  &\small{\ \ MC Result} &\small{\ \ Avg. Steps}  &\small{\ \ Num. of Iters}\\
\hline
AL &49.1\%& 165 & 2\\
\hline
Safest Policy  & 0.0\% & 8 & N.A.\\
\hline
$p^*=30\%$ & 17.2\%  & 121 & 10\\
\hline
$p^*=25\%$ & 9.3\%  & 136 & 14\\
\hline
$p^*=20\%$ & 17.2\% & 122 & 10\\
\hline
$p^*=15\%$ & 6.9\% & 118 & 22\\
\hline
$p^*=10\%$ & 7.2\% & 136 & 22\\
\hline
$p^*=5\%$ & 0.04\% & 83 & 50\\
\hline
\end{tabular}
\label{tab:cartpole1}
\end{center}
\end{table}
\vspace{-8mm}

We used $2000$ demonstrations for which the pole is held upright without violating any of the safety conditions for all 200 steps in each demonstration. The safest policy synthesized by PRISM-games is used as the initial safe policy. 
We also compare the different policies learned by CEGAL for different safety threshold $p^*$s. 
In Table~\ref{tab:cartpole1}, the policies are compared in terms of model checking results (`MC Result') on the PCTL property in (\ref{eq:spec1}) using the constructed MDP, the average steps (`Avg. Steps') that a policy (executed in the OpenAI environment) can hold across $5000$ rounds (the higher the better), and the number of iterations (`Num. of Iters) it takes for the algorithm to terminate (either converge to an $\epsilon$-close policy, or terminate due to $\sigma$, or terminate after $50$ iterations).
The policy in the first row is the result of using AL alone, which has the best performance but also a $49.1\%$ probability of violating the safety requirement.
The safest policy as shown in the second row is always safe has almost no performance at all. 
This policy simply letts the pole fall and thus does not risk moving the cart out of the range [-0.3, 0.3]. 
On the other hand, it is clear that the policies learnt using CEGAL always satisfy the safety requirement.
From $p^* = 30\%$ to $10\%$, the performance of the learnt policy is comparable to that of the AL policy. However, when the safety threshold becomes very low, e.g., $p^*=5\%$, the performance of the learnt policy drops significantly. This reflects the phenomenon that the tighter the safety condition is the less room for the agent to maneuver to achieve a good performance.

\subsection{Mountain-Car from OpenAI Gym}
\begin{figure}[h]
\centering
\subfigure[]{
	\hspace{0.15\linewidth}\begin{minipage}[c][0.7\width]{
	   0.25\textwidth}
	   \includegraphics[width=1.2\linewidth]{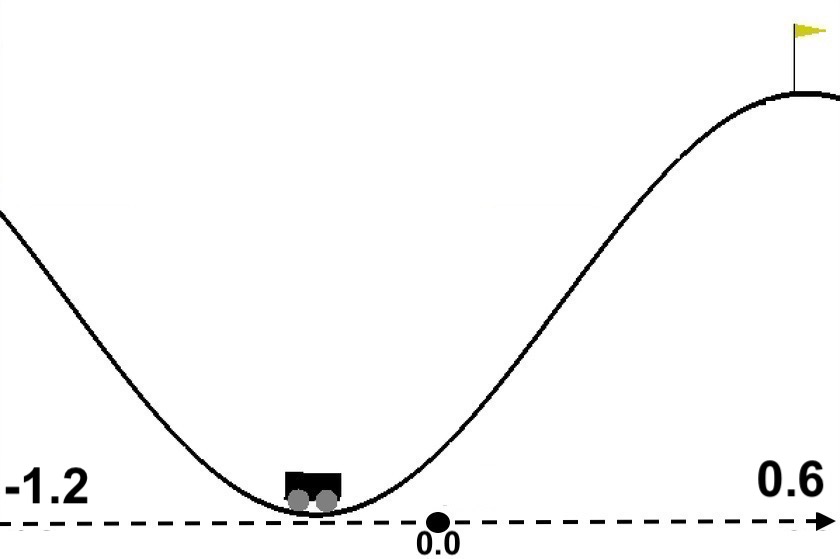}
	   \label{fig:mountaincar-v0}
	\end{minipage}\hspace{0.2\linewidth}
}
\subfigure[]{
	\begin{minipage}[c][0.7\width]{
	   0.25\textwidth}
	   \includegraphics[width=1.2\linewidth]{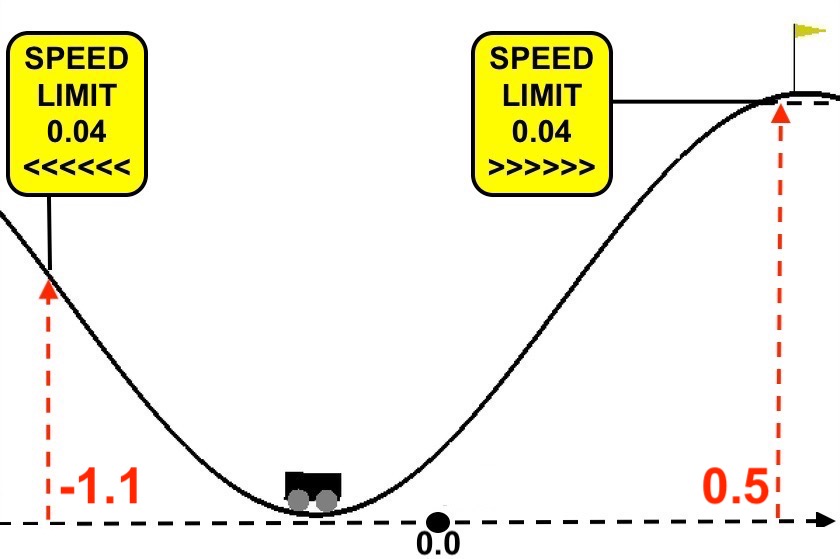}
	   \label{fig:mountaincar-v1}
	\end{minipage}\hspace{0.1\linewidth}
}
\vspace{-3mm} 
\caption{(a) The original mountain-car environment. (b) The mountain-car environment with traffic rules: when the distance from the car to the left edge or the right edge is shorter than $0.1$, the speed of the car should be lower than $0.04$.}  
\label{fig:mountaincar}
\end{figure}

Our third experiment uses the mountain-car environment from OpenAI Gym. As shown in Fig. {\ref{fig:mountaincar-v0}}, a car starts from the bottom of the valley and tries to reach the mountaintop on the right as quickly as possible. In each time step the car can perform one of the three actions, accelerating to the left, coasting, and accelerating to the right. The agent fails if the step length reaches the maximum ($t=66$). The velocity and position of the car are continuous values and observable while the exact dynamics are unknown\footnote{The MDP is built from sampled data. The feature vector for each state contains 2 exponential functions and 18 radial basis functions which respectively depend on the squared Euclidean distances between the current state and other 18 states which are uniformly distributed in the state space.}. In this game setting, the car cannot reach the right mountaintop by simply accelerating to the right. It has to accumulate momentum first by moving back and forth in the valley. The safety rules we enforce are shown in Fig.~{\ref{fig:mountaincar-v1}}. They correspond to speed limits when the car is close to the left mountaintop or to the right mountaintop (in case it is a cliff on the other side of the mountaintop). Similar to the previous experiments, we considered $2000$ expert demonstrations for which all of them successfully reach the right mountaintop without violating any of the safety conditions. 
The average number of steps for the expert to drive the car to the right mountaintop is $40$. 
We formalize the safety requirement in PCTL as (\ref{eq:spec2}). 
\vspace{-5mm} 
\vspace{-9mm}
\begin{equation}
\begin{split}
\Phi ::= P_{\leq p^*}[true\ \until^{\leq t}\ &(speed\leq -0.04\wedge position\leq-1.1)\\
&\vee(speed\geq 0.04\wedge position\geq0.5)]
\end{split}
\label{eq:spec2}
\end{equation}
\vspace{-2mm} 

We compare the different policies using the same set of categories as in the cart-pole example. The numbers are averaged over 5000 runs.
As shown in the first row, the policy learnt via AL\footnote{AL did not converge to an $\epsilon$-close policy in 50 iterations in this case.} has the highest probability of going over the speed limits. 
We observed that this policy made the car speed up all the way to the left mountaintop to maximize its potential energy. The safest policy corresponds to simply staying in the bottom of the valley. 
The policies learnt via CEGAL for safety threshold $p^*$ ranging from $60\%$ to $50\%$ not only have lower probability of violating the speed limits but also achieve comparable performance. As the safety threshold $p^*$ decreases further, the agent becomes more conservative and it takes more time for the car to finish the task. For $p^*=20\%$, the agent never succeeds in reaching the top within $66$ steps. 
\vspace{-8mm}
\begin{table}[hbt]
\begin{center}
\caption{In the mountain-car environment, {\it lower} average steps mean better performance. The safest policy is synthesized via PRISM-games.}
\begin{tabular}{|r|r|r|r|r|}
\hline 
\commentout{
\small{}  &\small{\ \ MC Result} &\small{\ \ Avg. Steps}  &\small{\ \ Unsafe Rate}&\small{\ \ Num. of Iters} \\ 
\hline \small
Policy Learnt via AL & 69.2\%& 54 & 100\% & 50\\
\hline
Safest Policy &  0.0\% &  $Fail$ & 0\% & N.A.\\
\hline
$p^*=60\%$&43.4\% & 57 & 33.2\% & 9\\
\hline
$p^*=50\%$ &46.9\%& 55 & 29.4\% & 23\\
\hline
$p^*=40\%$ &29.3\% & 61 & 0.6\% & 25\\
\hline
$p^*=30\%$ &18.9\% & 64 & 0.0\% & 17\\
\hline
$p^*=20\%$ &12.0\% & 66 & 3.5\% & 39\\
\hline
$p^*=10\%$ &7.6\% & $Fail$ & 0\%  & 40\\
}
\small{}  &\small{\ \ MC Result} &\small{\ \ Avg. Steps}  &\small{\ \ Num. of Iters} \\ 
\hline \small
Policy Learnt via AL & 69.2\%& 54 & 50\\
\hline
Safest Policy &  0.0\% &  $Fail$ & N.A.\\
\hline
$p^*=60\%$&43.4\% & 57 & 9\\
\hline
$p^*=50\%$ &47.2\%& 55 & 17\\
\hline
$p^*=40\%$ &29.3\% & 61 & 26\\
\hline
$p^*=30\%$ &18.9\% & 64 & 17\\
\hline
$p^*=20\%$ &4.9\% & $Fail$ & 40\\
\hline
\end{tabular}
\label{tab:mountaincar0}
\vspace{-12mm}
\end{center}
\end{table}

\section{Related Work}
\label{sec:related}
A taxonomy of AI safety problems is given in \cite{AmodeiOSCSM16} where the issues of misspecified objective or reward and insufficient or poorly curated training data are highlighted. There have been several attempts to address these issues from different angles. The problem of {\it safe exploration} is studied in \cite{moldovan2012safe} and \cite{DBLP:journals/corr/HeldMZSA17}. In particular, the latter work proposes to add a safety constraint, which is evaluated by amount of damage, to the optimization problem so that the optimal policy can maximize the return without violating the limit on the expected damage. An obvious shortcoming of this approach is that actual failures will have to occur to properly assess damage.

Formal methods have been applied to the problem of AI safety. 
In \cite{gillulay2011guaranteed}, the authors propose to combine machine learning and reachability analysis for dynamical models to achieve high performance and guarantee safety. In this work, we focus on probabilistic models which are natural in many modern machine learning methods. 
In \cite{DBLP:journals/corr/SadighKCSS14}, the authors propose to use formal specification to synthesize a control policy for reinforcement learning. They consider formal specifications captured in Linear Temporal Logic, whereas we consider PCTL which matches better with the underlying probabilistic model.  
Recently, the problem of {\it safe reinforcement learning} was explored in \cite{DBLP:journals/corr/abs-1708-08611} where a monitor (called shield) is used to enforce temporal logic properties either during the learning phase or execution phase of the reinforcement learning algorithm. The shield provides a list of safe actions each time the agent makes a decision so that the temporal property is preserved. 
In \cite{junges2016safety}, the authors also propose an approach for controller synthesis in reinforcement learning. In this case, an SMT-solver is used to find a scheduler (policy) for the synchronous product of an MDP and a DTMC so that it satisfies both a probabilistic reachability property and an expected cost property. 
Another approach that leverages PCTL model checking is proposed in \cite{mason2017assured}. A so-called abstract Markov decision process (AMDP) model of the environment is first built and PCTL model checking is then used to check the satisfiability of safety specification.
Our work is similar to these in spirit in the application of formal methods. However, while the concept of AL is closely related to reinforcement learning, an agent in the AL paradigm needs to learn a policy from demonstrations without knowing the reward function a priori.

A distinguishing characteristic of our method is the tight integration of formal verification with learning from data (apprenticeship learning in particular). Among imitation or apprenticeship learning methods, margin based algorithms \cite{Abbeel:2004:ALV:1015330.1015430}, \cite{Ng:2000:AIR:645529.657801}, \cite{Ratliff:2006:MMP:1143844.1143936} try to maximize the margin between the expert's policy and all learnt policies until the one with the smallest margin is produced. The apprenticeship learning algorithm proposed by Abbeel and Ng~\cite{Abbeel:2004:ALV:1015330.1015430} was largely motivated by the support vector machine (SVM) in that features of expert demonstration is maximally separately from all features of all other candidate policies. Our algorithm makes use of this observation when using counterexamples to steer the policy search process.
Recently, the idea of learning from failed demonstrations started to emerge. 
In \cite{shiarlis2016inverse}, the authors propose an IRL algorithm that can learn from both successful and failed demonstrations. It is done by reformulating maximum entropy algorithm in \cite{Ziebart:2008:MEI:1620270.1620297} to find a policy that maximally deviates from the failed demonstrations while approaching the successful ones as much as possible. However, this entropy-based method requires obtaining many failed demonstrations and can be very costly in practice. 

Finally, our approach is inspired by the work on formal inductive synthesis~\cite{jha-ai2017} and counterexample-guided inductive synthesis (CEGIS)~\cite{CEGIS}. These frameworks typically combine a constraint-based synthesizer with a verification oracle. In each iteration, the agent refines her hypothesis (i.e. generates a new candidate solution) based on counterexamples provided by the oracle. Our approach can be viewed as an extension of CEGIS where the objective is not just functional correctness but also meeting certain learning criteria. 

\vspace{-2mm}
\section{Conclusion and Future work}
\vspace{-1mm}
\label{sec:conclu}
We propose a counterexample-guided approach for combining probabilistic model checking with apprenticeship learning to ensure safety of the apprenticehsip learning outcome. Our approach makes novel use of counterexamples to steer the policy search process by reformulating the feature matching problem into a multi-objective optimization problem that additionally takes safety into account. Our experiments indicate that the proposed approach can guarantee safety and retain performance for a set of benchmarks including examples drawn from OpenAI Gym. In the future, we would like to explore other imitation or apprenticeship learning algorithms and extend our techniques to those settings.

\vspace{4mm}
\noindent
{\bf Acknowledgement.} This work is funded in part by the DARPA BRASS program under agreement number FA8750-16-C-0043 and NSF grant CCF-1646497.

\bibliographystyle{abbrv}
\bibliography{./references}
\end{document}